\documentclass[letterpaper]{article} 
\usepackage{aaai2026}  
\usepackage{times}  
\usepackage{helvet}  
\usepackage{courier}  
\usepackage[hyphens]{url}  
\usepackage{graphicx} 
\usepackage{natbib}  
\usepackage{caption} 
\usepackage{algorithm}
\usepackage{algorithmic}

\usepackage{newfloat}
\usepackage{listings}
\DeclareCaptionStyle{ruled}{labelfont=normalfont,labelsep=colon,strut=off} 
\floatstyle{ruled}
\newfloat{listing}{tb}{lst}{}
\floatname{listing}{Listing}
\usepackage{amsmath} 
\usepackage{amssymb}
\usepackage{booktabs}
\usepackage{multirow}
\usepackage{array}

\usepackage{times}
\usepackage{helvet}
\usepackage{courier}
\usepackage{xcolor}

\title{Focusing on Language: Revealing and Exploiting Language Attention Heads in Multilingual Large Language Models}
\author{
    Xin Liu\textsuperscript{\rm 1,\rm 2},
    Qiyang Song\textsuperscript{\rm 1,\rm 2}, 
    Qihang Zhou\textsuperscript{\rm 1,\rm 2}, 
    Haichao Du\textsuperscript{\rm 1,\rm 2}, 
    Shaowen Xu\textsuperscript{\rm 1,\rm 2}, \\
    Wenbo Jiang\textsuperscript{\rm 3}, 
    Weijuan Zhang\textsuperscript{\rm 1,\rm 2}, 
    Xiaoqi Jia\textsuperscript{\rm 1,\rm 2}\thanks{Corresponding Author.}
}
\affiliations{
    \textsuperscript{\rm 1}Institute of Information Engineering, Chinese Academy of Sciences\\
    \textsuperscript{\rm 2}School of Cyber Security, University of Chinese Academy of Sciences\\
    \textsuperscript{\rm 3}University of Electronic Science and Technology of China\\

    liuxin235@mails.ucas.ac.cn, jiaxiaoqi@iie.ac.cn
    
}

\usepackage{bibentry}

\begin{document}

\maketitle

\begin{abstract}
Large language models (LLMs) increasingly support multilingual understanding and generation. Meanwhile, efforts to interpret their internal mechanisms have emerged, offering insights to enhance multilingual performance. While multi-head self-attention (MHA) has proven critical in many areas, its role in multilingual capabilities remains underexplored. In this work, we study the contribution of MHA in supporting multilingual processing in LLMs. We propose Language Attention Head Importance Scores (LAHIS), an effective and efficient method that identifies attention head importance for multilingual capabilities via a single forward and backward pass through the LLM. Applying LAHIS to Aya-23-8B, Llama-3.2-3B, and Mistral-7B-v0.1, we reveal the existence of both language-specific and language-general heads. Language-specific heads enable cross-lingual attention transfer to guide the model toward target language contexts and mitigate off-target language generation issue, contributing to addressing challenges in multilingual LLMs. We also introduce a lightweight adaptation that learns a soft head mask to modulate attention outputs over language heads, requiring only 20 tunable parameters to improve XQuAD accuracy. Overall, our work enhances both the interpretability and multilingual capabilities of LLMs from the perspective of MHA.
\end{abstract}

\begin{links}
    \link{Code}{https://github.com/Linuxin-xxx/LAHIS}
\end{links}

\section{Introduction}
Large language models (LLMs) have substantially advanced natural language processing (NLP). With their increasing pretraining on multilingual corpora, these models now demonstrate impressive capabilities in understanding and generating text across diverse languages \citep{surveym, mllmsv}. As a result, enhancing and analyzing their multilingual abilities has emerged as a central research objective \citep{crosslft, crossconsistency, mlmpretrain}.

Alongside performance improvements, there is a growing interest in uncovering the internal mechanisms that enable multilingual processing in LLMs. Gaining a deeper understanding of these mechanisms can facilitate the development of more interpretable and controllable models. Recent studies have made progress on this area by identifying language-specific neurons or analyzing the function of each layer \citep{PLND, LAPE, unveiling, lostml, beyonden, llamalan}. However, one fundamental component of the Transformer architecture, multi-head self-attention (MHA), remains relatively underexplored in this context.

In other domains, studies have identified attention heads that perform specialized functions, such as induction heads~\citep{inductionhead}, retrieval heads~\citep{retrievalhead}, successor heads~\citep{successorhead}, and safety heads~\citep{safetyrole}. These roles indicate that attention heads can exhibit functional specialization. Motivated by this, we hypothesize that multilingual LLMs may also contain attention heads that are selectively specialized for processing particular languages.

In this paper, we investigate the contribution of individual attention heads within MHA to the multilingual capabilities of LLMs. To this end, we propose \textbf{Language Attention Head Importance Scores (LAHIS)}, an efficient method for attributing head importance with respect to multilingual performance. LAHIS utilizes a soft mask, implemented as a trainable matrix shaped to match the model’s attention structure, to estimate head importance by tracking its updates during training. This process requires only a single forward and backward pass over a target-language corpus.

By applying LAHIS to Aya-23-8B, Llama-3.2-3B, and Mistral-7B-v0.1, we obtain attention head importance matrices across different languages and identify \textbf{language heads} that contribute significantly to multilingual capabilities. We define those that are highly important for specific languages as \textbf{language-specific heads}, and empirically validate both their effectiveness and specialization. Furthermore, we identify a set of language-general heads that consistently receive high importance scores across all languages.

Language-specific heads can help address challenges faced by multilingual LLMs. In real-world applications such as dialogue systems and retrieval-augmented generation, multilingual input contexts are common. However, when conflicting facts are presented in different languages, the model’s information selection can become unpredictable and biased \citep{ragmlprefer}. Our experiments demonstrate that by enhancing or suppressing language-specific heads, we can steer the model’s attention and shift its preference toward the target-language context, thereby enabling more reliable and controllable cross-lingual reasoning.

We further demonstrate that modulating language-specific heads provides a practical mechanism to mitigate off-target language generation \citep{offtargettranslation}, a common issue where multilingual models are likely to output high-resource languages like English, even when the input is in a different language. By suppressing English-heads during generation, we effectively realign the output language with the input. Our experiments on the XL-Sum dataset show that this targeted intervention restores language fidelity while maintaining high summarization quality, revealing a controllable axis for correcting multilingual generation asymmetries in LLMs.

Moreover, building on the discovery of language heads, we propose a lightweight adaptation method that enhances multilingual performance by training a small matrix aligned with the attention structure of the LLM, without modifying the LLM’s original weights. This trainable matrix updates only a few parameters associated with the identified heads and is inserted into the LLM as a gate to modulate the corresponding attention outputs during training or inference. Experiments on the XQuAD dataset across three models show that this adaptation, tuning only 14–20 parameters, yields an average accuracy improvement of 5 percentage points over the vanilla models, and outperforms the random-heads baseline by 4 points.

Our key contributions are as follows:
\begin{enumerate}  
\item We propose LAHIS, a lightweight and efficient method for estimating attention head importance with respect to multilingual capabilities. Applied to three LLMs across diverse languages, LAHIS reveals the existence of language-specific heads critical to individual language performance, and language-general heads beneficial across languages.
\item We show that language-specific heads offer a mechanism to control model behavior in multilingual contexts. By enhancing or suppressing these heads, we steer the model's attention toward the target language, enhancing cross-lingual reasoning and reducing bias from conflicting multilingual inputs. 
\item We demonstrate that modulating language-specific heads effectively mitigates off-target language generation. Our experiments show that suppressing English-heads realigns the outputs with input languages while maintaining generation quality.
\item We propose a lightweight adaptation that enhances multilingual performance by inserting a simple, attention-aligned trainable matrix to modulate certain language heads outputs. Tuning only 14–20 parameters, it improves XQuAD accuracy by 5 points on average and surpasses the random-heads baseline by 4 points.
\end{enumerate}

\section{Method}
\subsection{Preliminary}
In this section, we briefly introduce the Transformer architecture and the multi-head self-attention (MHA) mechanism \citep{transformer}, as they form the foundation for the rest of the paper. We mainly focus on the decoder-only transformer architecture, which is widely adopted in modern LLMs. This architecture generally consists of multiple layers, each comprising a multi-head self-attention module and a feed-forward network.

The self-attention mechanism computes dependencies between tokens in an input sequence, assigning different weights to enable the LLMs to focus on the most relevant information. Multi-head self-attention further splits internal representations into multiple subspaces, allowing the models to capture diverse semantic features that a single head may struggle to represent effectively \citep{transformer}.

Each head in the MHA module is implemented as a Scaled Dot-Product Attention unit. The $i$-th head, $\mathrm{head}_i$, is calculated as:

\begin{equation}\small
\mathrm{head}_i = \mathrm{Attention}(Q_i, K_i, V_i) = \mathrm{softmax}\left( \frac{Q_i K_i^T}{\sqrt{d_k}} \right)V_i
\end{equation}

The inputs consist of queries $Q_i$, keys $K_i$, and values $V_i$, each with a dimensionality of $d_k$. In self-attention, $Q_i$, $K_i$, and $V_i$ are obtained from the same internal representation produced by the previous layer, through different linear projections. In this way, each head performs its computation independently. The outputs of all heads $\mathcal{H}$ are then concatenated and projected by a parameter matrix $W_O$ to form the final output of the MHA module:

\begin{equation}\small
\mathrm{MultiHead} = \mathrm{Concat}(\mathrm{head}_1, \ldots, \mathrm{head}_h)W_O
\end{equation}

Moreover, several models, such as Llama-3, adopt the Grouped-Query Attention (GQA) mechanism \citep{GQA}, in which multiple queries within a group share the same keys ($K$) and values ($V$). However, this shared use of $K$ and $V$ does not eliminate head-wise differentiation, as each head still employs its own distinct queries ($Q$).

Specifically, for layer $l$ in an LLM, given the internal representation $\mathcal R^{l-1}$ from the previous layer and the trainable parameters $W^l_Q$, $W^l_K$, and $W^l_V$ of the current layer, attention head $\mathrm{head}^l_i$ is computed as:

\begin{equation}\small
\mathrm{head}^l_i = \mathrm{Attention}(\mathcal R^{l-1}W^l_Q, \mathcal R^{l-1}W^l_K, \mathcal R^{l-1}W^l_V)
\end{equation}

In this paper, a gate parameter $g_i$ is used to control the magnitude of each attention head $head_i$:
\begin{equation}\small
\tilde{head}_i = g_i \cdot head_i, \text{where } g_i \in \mathbb{R}_{\ge 0} \text{ and }
g_i \begin{cases}
> 1 & \text{(enhance)} \\
\in [0, 1) & \text{(weaken)} \\
= 0 & \text{(deactivate)}
\end{cases}
\end{equation}

\subsection{Language Attention Head Importance Scores}
\label{sec:LAHIS}
In this paper, we focus on the attention heads within multilingual LLMs. We propose a lightweight and efficient method, Language Attention Head Importance Scores (\textbf{LAHIS}), which estimates the importance of each attention head for multilingual capability. Specifically, it computes an importance matrix that reflects how crucial each attention head is to a specific language, using only a single forward and backward pass of the LLM on a corpus in that language. This is achieved by a trainable matrix designed to efficiently approximate each head’s contribution.

Following prior works \citep{safetyrole, sixteenheads, LAPE, PLND}, we estimate the importance of an attention head $h$ by measuring the change in model loss $\Delta \mathcal{L}$ before and after disabling it via a masking mechanism. Given a language-specific corpus $\mathcal{X}_c$ in language $c$, the model computes the change in loss $\mathcal{L}(x)$ when an attention head is deactivated. As shown in earlier studies \citep{pruned2019, sixteenheads}, many attention heads are redundant for specific tasks and can be pruned with minimal degradation in performance. Hence, if disabling a particular head leads to a significant increase in loss, it indicates that the head plays an important role in processing language $c$.

Formally, for an LLM with $n_l$ layers and $n_h$ heads per layer, we construct the attention head importance matrix $\mathrm{ImpScore}_c \in \mathbb{R}^{n_l \times n_h}$ for language $c$, where each entry is defined based on the loss difference $\Delta \mathcal{L}$ when deactivating a specific attention head. Specifically, the importance score for attention head $h_i$ is given by:

\begin{equation}\small
\mathrm{ImpScore}_c(h_i) = \mathbb{E}_{x_c \in \mathcal{X}_c}\left[ \mathcal{L}(h_i{=}0 \mid x_c) - \mathcal{L}(x_c) \right]
\end{equation}

with $h_i \in \mathcal{H}$ indexing all attention heads.

However, explicitly disabling each attention head one by one is computationally expensive, especially for large-scale models such as Aya-23-8B, which contains 1024 attention heads. To address this, inspired by previous works \citep{sixteenheads, taylorprune}, we approximate the loss difference $\Delta \mathcal{L}$ using a \textit{first-order Taylor expansion}. Specifically, we introduce a trainable matrix $\mathcal{M} \in \mathbb{R}^{n_l \times n_h}$, referred to as the soft attention head mask, to efficiently estimate the head-wise importance through a single forward and backward pass. The approximated loss difference $\Delta \tilde{\mathcal{L}}$ is given by:

\begin{equation}\small
\Delta \tilde{\mathcal{L}} = \mathbb{E}_{x_c \in \mathcal{X}_c} \left[ \left| m_i \cdot \frac{\partial \mathcal{L}(x_c)}{\partial m_i} \right| \right]
\end{equation}

where $m_i \in \mathcal{M}$ denotes the mask parameter corresponding to attention head $h_i$.

Furthermore, since we aim to identify heads whose removal leads to an increase in loss, we compute the proportion of negative gradients with respect to each mask parameter, denoted by $W_{\text{neg}}$:

\begin{equation}\small
W_{\text{neg}} = \mathbb{E}_{x_c \in \mathcal{X}_c} \left[ \mathbb{I} \left( \frac{\partial \mathcal{L}(x_c)}{\partial m_i} < 0 \right) \right]
\end{equation}


Finally, we define the refined attention head importance matrix $\mathrm{LAHIS}_c \in \mathbb{R}^{n_l \times n_h}$ for language $c$ by weighting the loss difference $\Delta \mathcal{L}$ with the factor $W_{\text{neg}}$ that emphasizes heads with beneficial gradients. Formally, each importance score of $\mathrm{LAHIS}_c$ is computed as:

\begin{equation}\small
\label{eq:importance_final}
\mathrm{LAHIS}_{c}(h_i) = \mathbb{E}{x_c \in \mathcal{X}_c} \left[ \left| m_i \cdot \frac{\partial \mathcal{L}(x_c)}{\partial m_i} \right| \cdot \mathbb{I}\left( \frac{\partial \mathcal{L}(x_c)}{\partial m_i} < 0 \right) \right]
\end{equation}

In summary, each score in the attention head importance matrix reflects the degree to which a given head contributes to the model’s ability to handle a specific language. Attention heads with higher scores are deemed more critical and can be regarded as language-specific heads that play a specialized role in multilingual processing.

\section{Experiments}
\subsection{Experiment Setting}
\textbf{Models} We conduct experiments on three widely-used open-source multilingual models: Aya-23-8B \citep{aya23}, Llama-3.2-3B \citep{llama3}, and Mistral-7B-v0.1 \citep{mistral7b}. Aya-23-8B is particularly optimized for multilinguality and supports 23 languages, including both high-resource and low-resource languages. Llama-3.2-3B officially supports eight languages, while it has been trained on a broader collection of languages beyond these supported ones. Mistral-7B-v0.1 is also pretrained on a mixture of multilingual data and exhibits multilingual capabilities.

\textbf{Datasets} For language modeling, we use perplexity (PPL) on the Wikipedia dataset, which contains cleaned articles across multiple languages, with one subset per language. For multilingual summarization, we use XL-Sum~\citep{xlsum}, a large-scale dataset of article-summary pairs in 44 languages from BBC, covering both low and high-resource languages. Summary quality is measured using BERTScore F1 \citep{bertscore}. For multilingual understanding, we adopt XQuAD~\citep{xquad}, a closed-domain QA benchmark in ten languages that evaluates cross-lingual transfer and comprehension.

\textbf{Languages} Based on the languages supported by the selected models and datasets, our experiments are conducted on the following languages: English (\texttt{en}), Spanish (\texttt{es}), Italian (\texttt{it}), German (\texttt{de}), Portuguese (\texttt{pt}), Greek (\texttt{el}), Chinese (\texttt{zh}), Japanese (\texttt{ja}), Korean (\texttt{ko}), Thai (\texttt{th}), Hindi (\texttt{hi}), Vietnamese (\texttt{vi}), and Indonesian (\texttt{id}). These languages cover both high-resource and low-resource settings to ensure diversity in our evaluation.

\subsection{Obtaining Attention Head Importance Matrix}
\label{sec:imp_matrix}
We apply LAHIS to obtain attention head importance matrices for multilingual capabilities in Aya-23-8B, Llama-3.2-3B, and Mistral-7B-v0.1, focusing on the following language sets for each model respectively: \texttt{(en, zh, hi, vi, es, pt, id, ja, ko, el)}, \texttt{(en, pt, it, de, th, id, vi, ja, ko, el)} and \texttt{(en, es, vi, hi, ja, th, el)}. As described in \S~\ref{sec:LAHIS}, we perform forward and backward passes over each language's Wikipedia corpus, which contains hundreds of thousands of tokens, to compute attention head importance scores using Equation~\ref{eq:importance_final}.

\begin{figure}[t]
\centering
\includegraphics[width=\columnwidth]{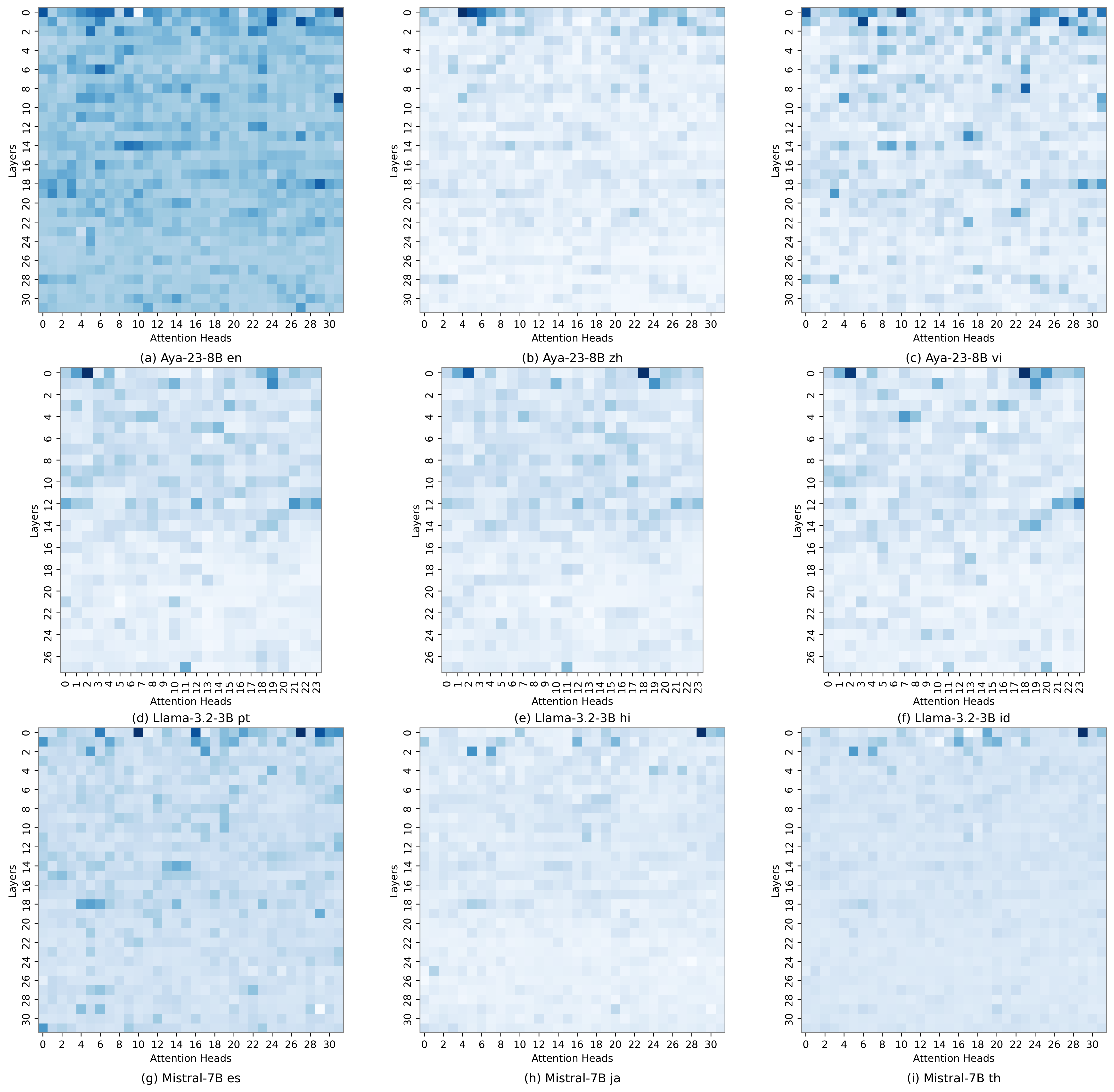}
\caption{Illustration of attention head importance matrices obtained by LAHIS. Darker cells indicate higher importance scores. Most cells are light, showing that only a small subset of heads are highly important.}
\label{fig:importance_matrices}
\end{figure}

Figure~\ref{fig:importance_matrices} presents the attention head importance matrices as heatmaps, where darker regions represent higher importance scores. Each heatmap is primarily composed of lighter areas with a few dark blocks, indicating that most attention heads contribute little to the model's performance on the target language, while a small subset has a disproportionately high impact. These high-importance heads are likely to capture language-specific patterns.

Furthermore, the identified high-importance heads are predominantly located in the lower layers of the LLMs, where attention heads exhibit greater feature differentiation \citep{pruned2019}.

\subsection{Language-General Heads}
Certain attention heads show consistently high importance across all languages, suggesting a key role in general language processing. To validate this, we disable these heads and evaluate LLMs on the XL-Sum dataset (500 samples per language) using BERTScore F1, which reflects both language understanding and generation abilities. Mistral-7B is excluded due to its tendency to generate English summaries for non-English inputs, resulting in artificially low evaluation scores. For Aya-23-8B and Llama-3.2-3B, we select the top 4\% and 19\% most important heads per language, respectively, and identify the ones shared across all languages, which account for approximately 1\% and 5\% of total heads. 

As shown in Table~\ref{tab:xlsum}, disabling these heads consistently degrades summarization quality across languages on two models (-12.3 and -19.2 F1 score on average), while randomly removing the same number of heads has minimal impact. In some cases, disabling these heads causes the models to produce repetitive or nonsensical outputs, indicating severe damage to their language capabilities. Therefore, we argue that these heads are universally important across languages and refer to them as language-general heads.

\begin{table}[t]\small
\setlength\tabcolsep{5pt}
\centering
\begin{tabular}{lcccccccc}
\toprule
\multicolumn{9}{c}{Aya-23-8B} \\ 
\midrule
 & zh & hi & vi & es & pt & id & ko & \textbf{Avg} \\ 
\cmidrule(lr){2-9}
VM & 89.1 & 85.7 & 79.3 & 69.6 & 72.7 & 68.3 & 84.6 & 78.5 \\
RH & 88.5 & 86.5 & 77.8 & 66.8 & 72.9 & 67.0 & 84.8 & 77.7 \\
\textbf{GH} & 72.0 & 84.0 & 69.0 & 58.4 & 63.5 & 47.9 & 69.0 & \textbf{66.2} \\ 
\midrule
\multicolumn{9}{c}{Llama-3.2-3B} \\ 
\midrule
 & en & pt & th & id & vi & ja & ko & \textbf{Avg} \\ 
\cmidrule(lr){2-9}
VM & 56.2 & 59.1 & 56.3 & 66.5 & 71.4 & 72.2 & 82.1 & 66.3 \\
RH & 54.8 & 61.2 & 55.2 & 63.7 & 75.5 & 74.2 & 82.8 & 66.8 \\
\textbf{GH} & 48.0 & 53.9 & 39.0 & 46.9 & 38.7 & 61.4 & 59.0 & \textbf{49.6}\\ \bottomrule
\end{tabular}
\caption{Impact of deactivating language-general heads on XL-Sum. Disabling these heads significantly reduces the average F1 scores ($\uparrow$), whereas randomly removing heads has negligible impact. \textit{VM: Vanilla Model; RH: Random Heads Deactivated; GH: Language-General Heads Deactivated.}}
\label{tab:xlsum}
\end{table}

\subsection{Language-Specific Heads}
\S~\ref{sec:imp_matrix} reveals that a small subset of attention heads plays a dominant role in supporting individual language capabilities in multilingual LLMs. We refer to these as language-specific heads. Specifically, based on the attention head importance matrices obtained by LAHIS, we select the top 2\% highest-scoring heads per language, excluding language-general heads to ensure specificity. We then evaluate both the effectiveness and the language specificity of these heads. 

To evaluate effectiveness, we deactivate the identified language-specific heads and measure the impact on perplexity (PPL) using the multilingual Wikipedia corpus with several hundred thousand tokens per language. An increase in PPL indicates performance degradation in language ability, reflecting the importance of the deactivated heads. As a baseline, we randomly deactivate the same number of heads. As shown in Table~\ref{tab:ppl_impact}, disabling language-specific heads consistently leads to larger PPL increases than both the vanilla model and the random baseline, demonstrating the effectiveness of the language-specific heads identified by LAHIS.

\begin{figure}[t]
\centering
\includegraphics[width=\columnwidth]{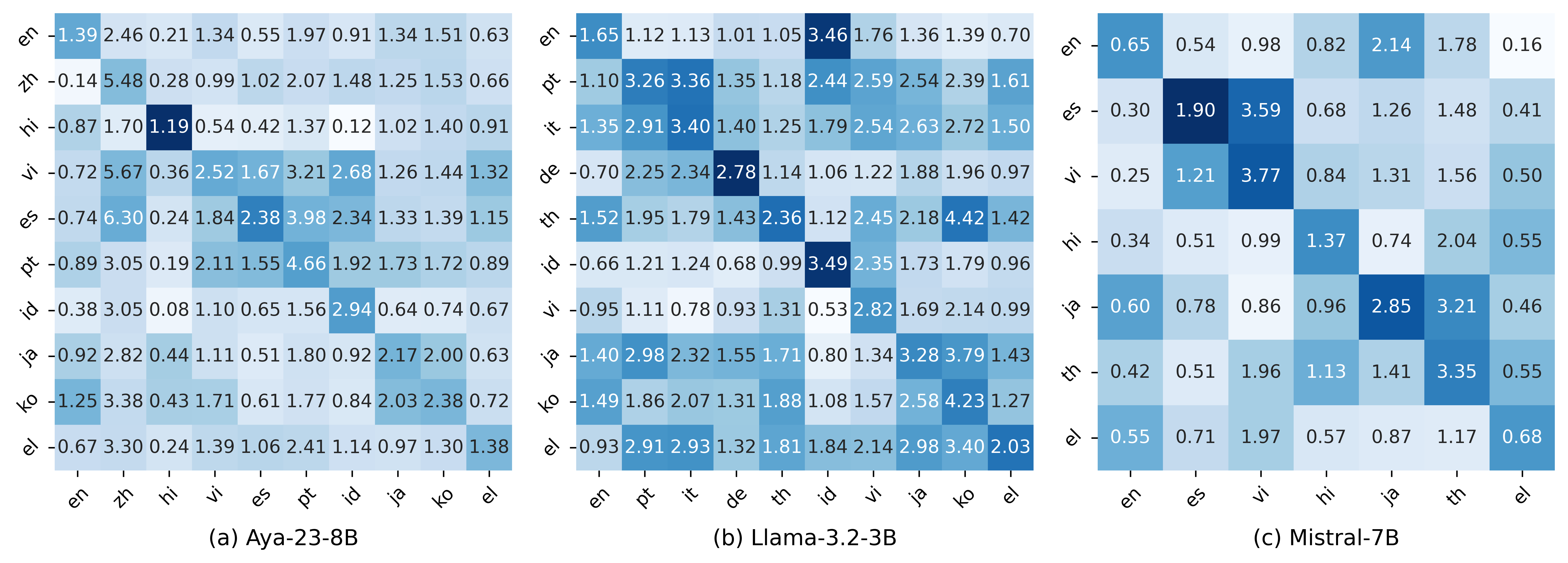}
\caption{PPL impact from deactivating language-specific heads. Cell $(i,j)$ shows the PPL increase on language $j$ when disabling language $i$’s heads. The dark diagonals indicate the the specificity of heads identified by LAHIS.}
\label{fig:head_specificity_heatmap}
\end{figure}

To assess specificity, we evaluate whether each set of language-specific heads primarily affects its corresponding language by examining the cross-lingual impact of deactivating them. Figure~\ref{fig:head_specificity_heatmap} presents the resulting PPL increases, where each cell $(i,j)$ denotes the PPL change on language $j$ when the heads specific to language $i$ are disabled. The consistently darker diagonal entries indicate that almost every language is affected the most by the deactivation of its own heads, confirming their specificity.

\begin{table}[t]\small
\centering
\begin{tabular}{lccccc}
\toprule
\multicolumn{6}{c}{Aya-23-8B} \\ 
\midrule
 & en & zh & hi & vi & es \\ 
\cmidrule(lr){2-6}
Vanilla Model & 8.18 & 8.66 & 2.89 & 6.38 & 4.22 \\
Random Heads & 8.26 & 8.94 & 2.92 & 6.42 & 4.36 \\
\textbf{Language Heads} & 9.57 & 14.1 & 4.08 & 8.90 & 6.60 \\ 
\cmidrule(lr){2-6}
 & pt & id & ja & ko & el \\ 
\cmidrule(lr){2-6}
Vanilla Model & 11.2 & 7.20 & 8.57 & 7.24 & 6.66 \\
Random Heads & 11.5 & 7.34 & 8.75 & 7.47 & 6.78 \\
\textbf{Language Heads} & 15.8 & 10.1 & 10.7 & 9.62 & 8.04 \\ 
\midrule
\multicolumn{6}{c}{Llama-3.2-3B} \\ 
\midrule
 & en & pt & it & de & th \\ 
\cmidrule(lr){2-6}
Vanilla Model & 8.36 & 9.09 & 8.76 & 6.26 & 6.35 \\
Random Heads & 8.65 & 9.28 & 8.84 & 6.47 & 6.63 \\
\textbf{Language Heads} & 10.01 & 12.35 & 12.16 & 9.04 & 8.71 \\ 
\cmidrule(lr){2-6}
 & id & vi & ja & ko & el \\ 
\cmidrule(lr){2-6}
Vanilla Model & 2.98 & 8.93 & 9.04 & 10.92 & 4.78 \\
Random Heads & 3.00 & 9.13 & 9.41 & 11.15 & 4.89 \\
\textbf{Language Heads} & 6.47 & 11.75 & 12.32 & 15.15 & 6.81 \\ 
\midrule
\multicolumn{6}{c}{Mistral-7B} \\ 
\midrule
 & en & es & vi & hi & ja \\ 
\cmidrule(lr){2-6}
Vanilla Model & 5.08 & 3.26 & 5.14 & 3.43 & 6.96 \\
Random Heads & 5.17 & 3.32 & 5.20 & 3.46 & 6.99 \\
\textbf{Language Heads} & 5.73 & 5.16 & 8.91 & 4.80 & 9.81 \\ 
\cmidrule(lr){2-6}
 & th & el & - & - & - \\ 
\cmidrule(lr){2-6}
Vanilla Model & 4.35 & 2.73 & - & - & - \\
Random Heads & 4.36 & 2.76 & - & - & - \\
\textbf{Language Heads} & 7.70 & 3.41 & - & - & - \\ 
\bottomrule
\end{tabular}
\caption{Impact of deactivating language-specific heads on PPL ($\downarrow$). Disabling language-specific heads significantly increases PPL compared to both the vanilla models and random heads removal, highlighting the effectiveness of the LAHIS method in identifying language-specific heads.}
\label{tab:ppl_impact}
\end{table}

\subsection{Transferring Cross-Lingual Attention}
In interactive dialogue or retrieval-augmented generation (RAG) scenarios, input contexts are often multilingual \citep{ragml, ragmlsetting}. However, when cross-lingual information conflicts, they exhibit a preference for the language of the context during information extraction, driven by prompt design or input-language factors \citep{ragmlprefer, ragmlsetting, faux}, potentially misaligning with users’ task-specific language requirements.

In this section, we investigate whether language-specific heads can steer the model’s attention across languages. Specifically, we test whether enhancing or weakening them can guide the model’s attention toward a target language.

\begin{figure}[t]
\centering
\includegraphics[width=0.95\columnwidth]{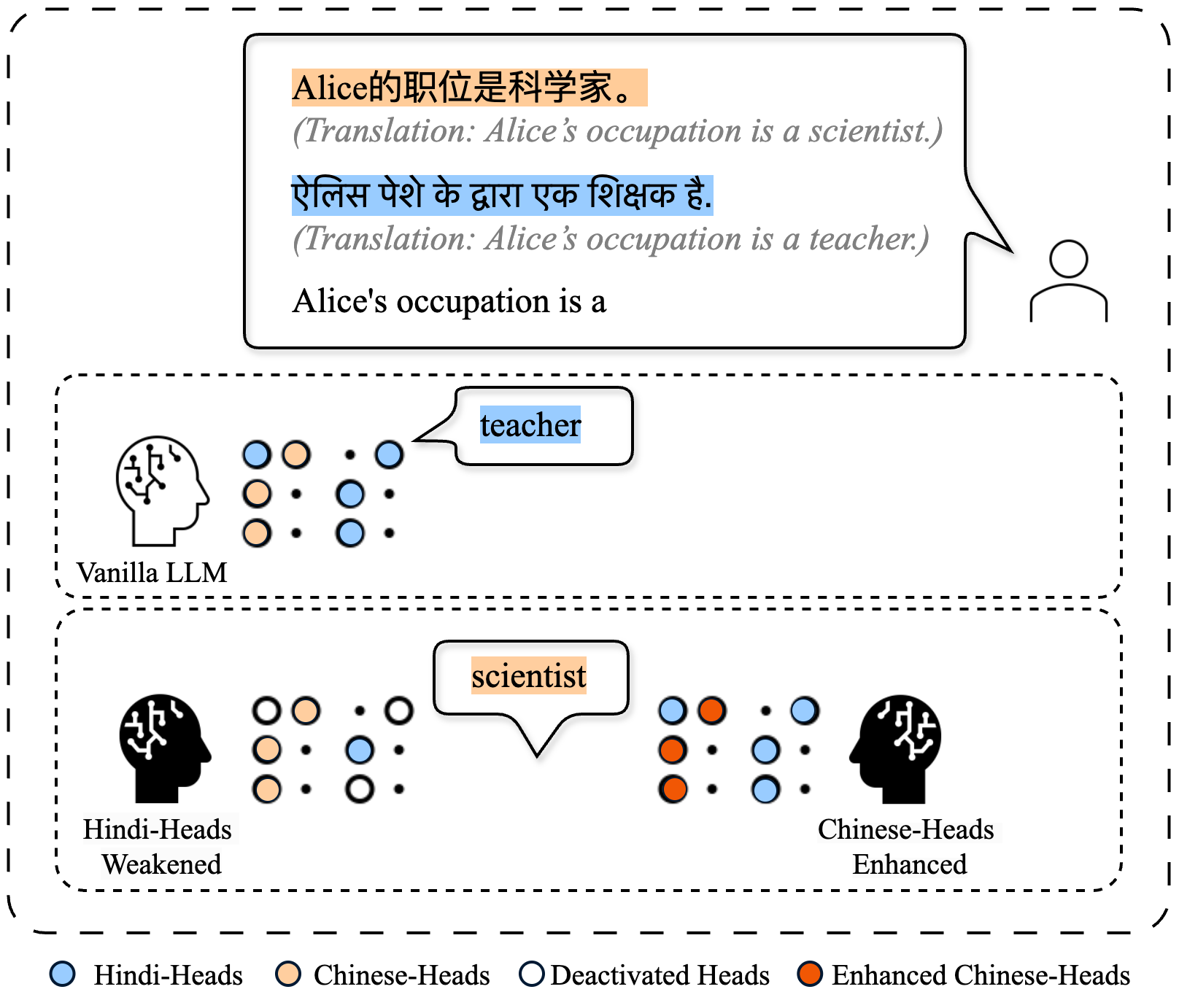}
\caption{An illustration of shifting LLM’s multilingual attention via head intervention. The vanilla model chooses "teacher" from the Hindi context, but switches to "scientist" from the Chinese context after weakening Hindi-heads or enhancing Chinese-heads.}
\label{fig:ta_demo}
\end{figure}

\begin{figure}[t]
\centering
\includegraphics[width=\columnwidth]{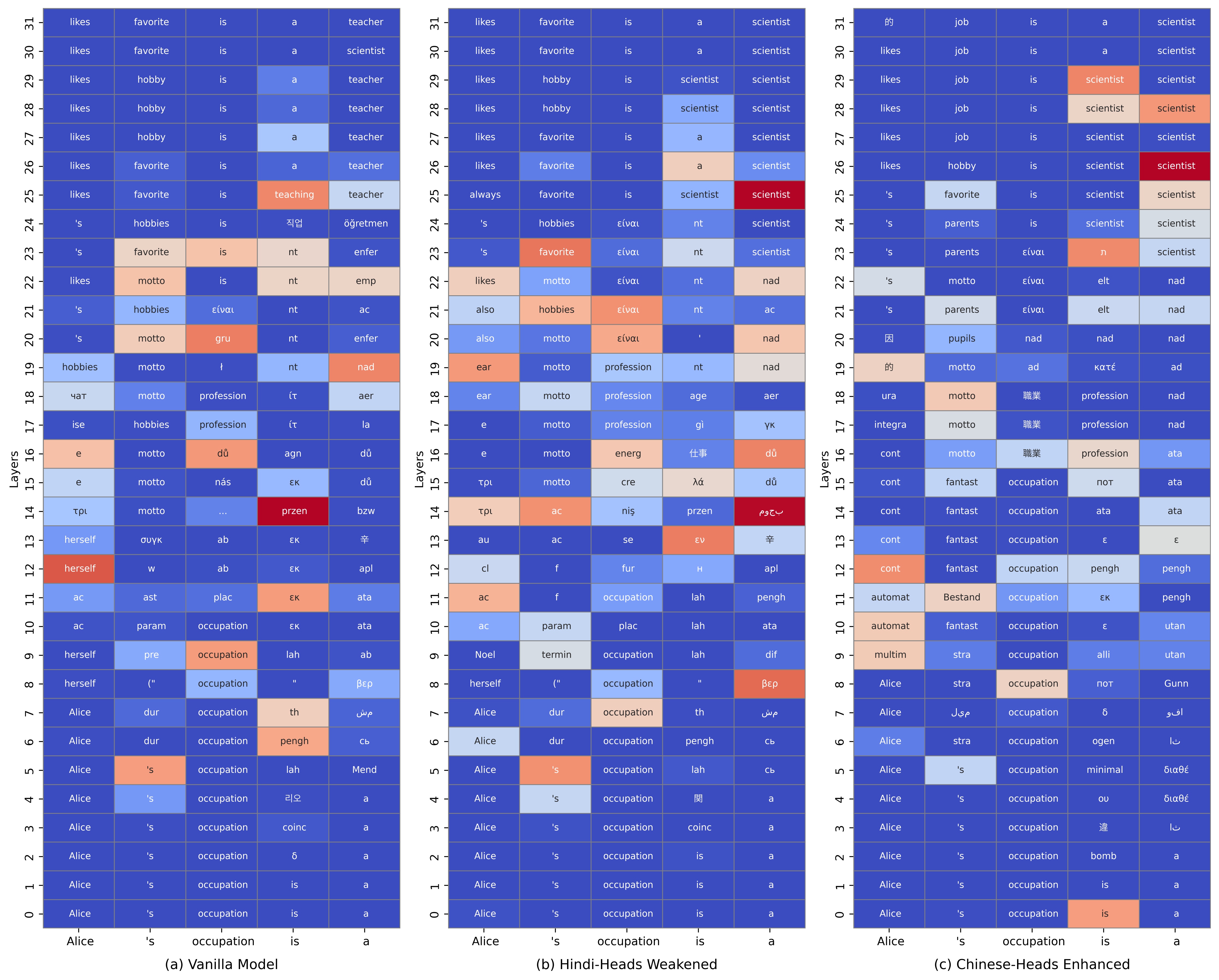}
\caption{Layer-wise vocabulary projections of Aya-23-8B illustrating the multilingual attention shift.}
\label{fig:aya_ta}
\end{figure}

As illustrated in Figure~\ref{fig:ta_demo}, the input provides conflicting facts about Alice’s occupation: “scientist” in Chinese and “teacher” in Hindi. When queried in English, the model initially answers “teacher”, reflecting the Hindi input. Enhancing Chinese-heads or suppressing Hindi-heads shifts the answer to “scientist”, aligning with the Chinese context. Figure~\ref{fig:aya_ta} further visualizes vocabulary-projected activations across Aya-23-8B’s layers using LogitLens~\citep{logitlens}, showing that manipulating language-specific heads effectively redirects attention to the intended language, altering the final output.

To verify cross-lingual attention transfer via language-specific heads, we design prompts with conflicting descriptions of Alice’s profession in two languages (\texttt{Context1} in language A and \texttt{Context2} in language B) and an English question. The model’s predicted profession indicates its preference for one language’s context. We then enhance language A’s heads or deactivate language B’s heads to examine whether the output shifts toward \texttt{Context1}.

\begin{figure}[t]
\centering
\includegraphics[width=\columnwidth]{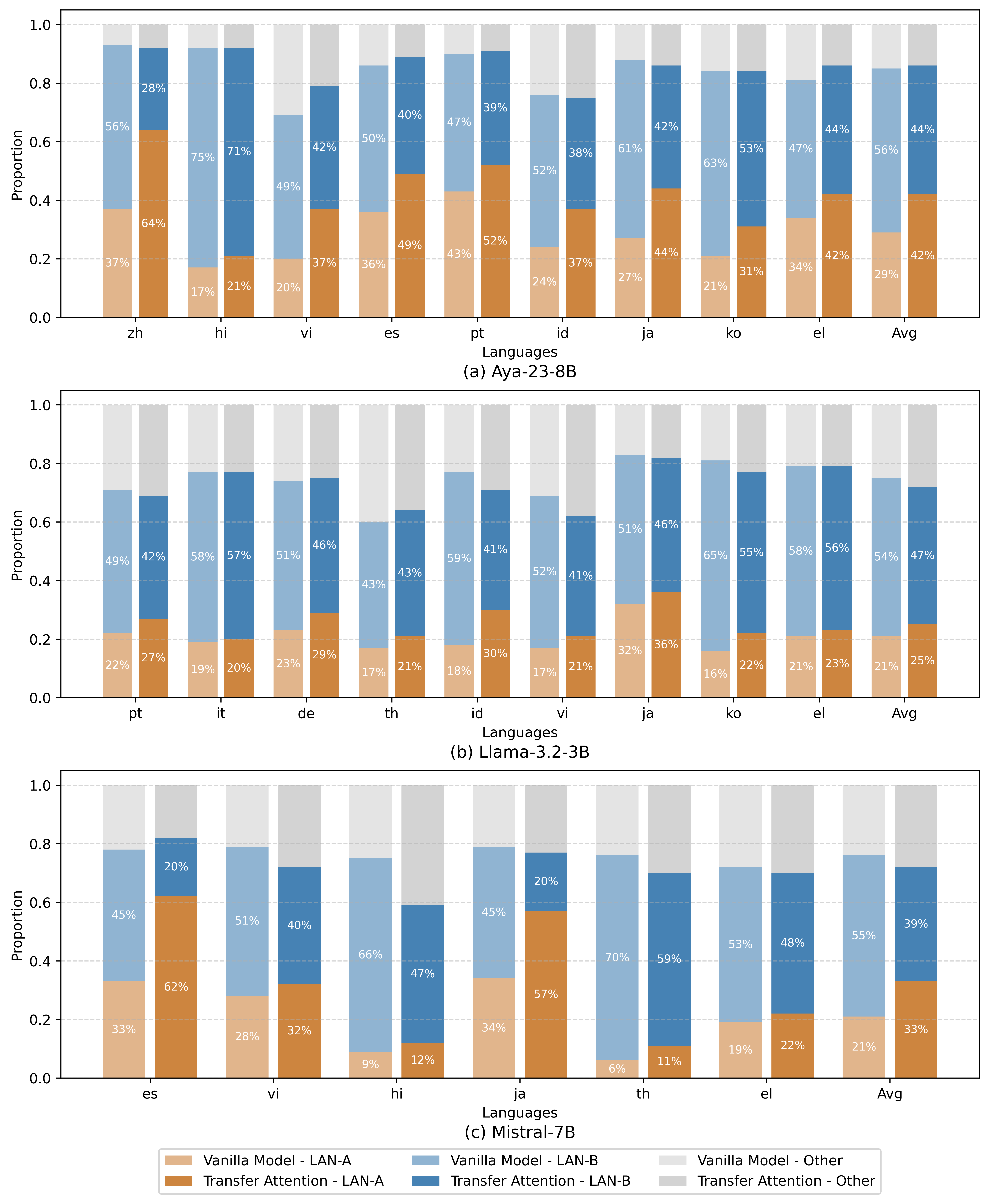}
\caption{Influence of language-specific heads on multilingual prediction shifting. Given prompts with conflicting facts in two languages, enhancing heads for language A or disabling heads for B increases the model’s reliance on language A’s context, indicating that language-specific heads can steer cross-lingual attention and influence predictions.}
\label{fig:ta_results}
\end{figure}

Figure~\ref{fig:ta_results} shows experimental results on Aya-23-8B, Llama-3.2-3B, and Mistral-7B with 450–720 samples per language. Models initially prefer \texttt{Context2}, likely due to its proximity to the question. After intervention, preference for \texttt{Context1} rises by about 10 percentage points, while reliance on \texttt{Context2} drops by 12 points. These results confirm that language-specific heads enable effective cross-lingual attention transfer.

\subsection{Mitigating Off-Target Language Generation}
Multilingual LLMs exhibit \textit{off-target language generation} \citep{LAPE, offtargettranslation, ontheofftarget, improvedzeroshot}, where outputs do not match the input language. This misalignment adversely affects tasks requiring language consistency. This phenomenon is also observed in our experiments on XL-Sum dataset with Mistral-7B, that the model generates English summaries despite non-English inputs and prompts, resulting in reduced evaluation scores. The bias toward English is primarily attributed to its dominance in the pretraining corpus, leading the model to develop stronger proficiency and preference for English \citep{dollmspeak}.

\begin{figure}[t]
\centering
\includegraphics[width=\columnwidth]{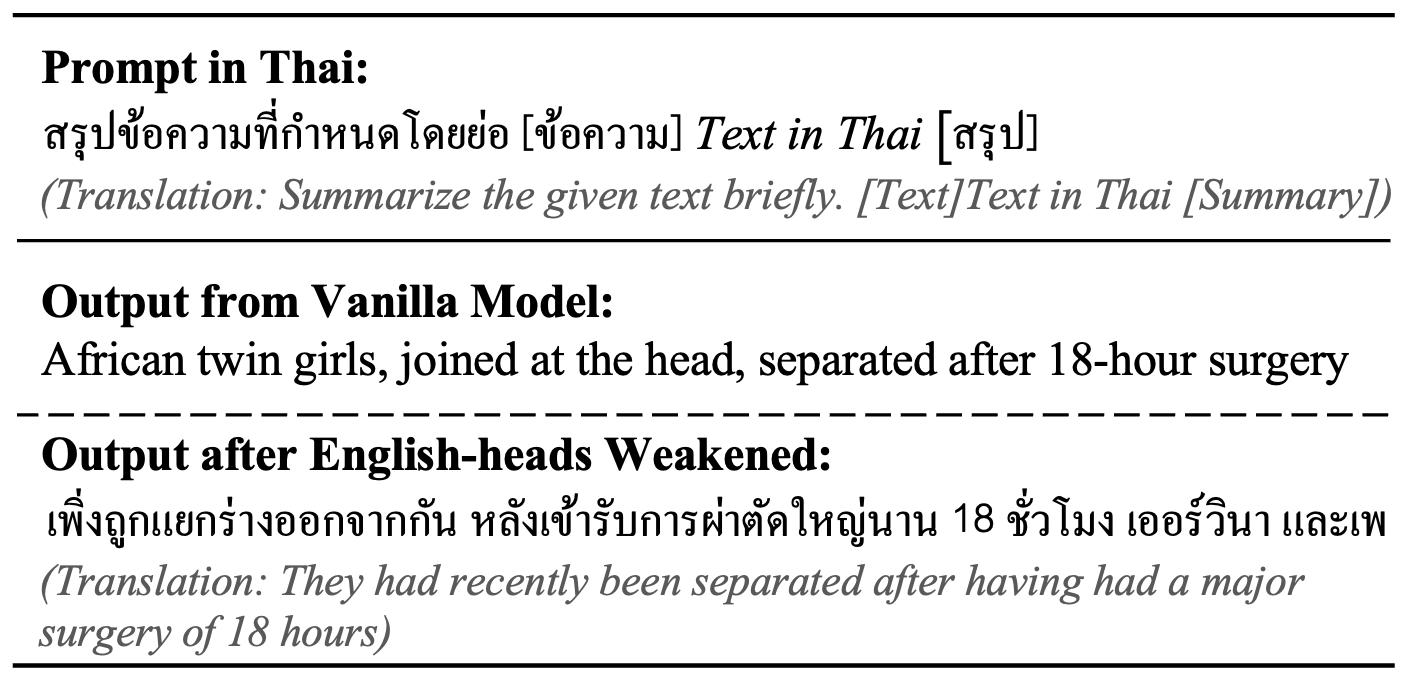}
\caption{An example of off-target language generation and mitigation. The LLM summarizes Thai input in English but outputs correctly after weakening English-heads.}
\label{fig:thaidemo}
\end{figure}

To address this issue, we reduce the model’s attention to English to encourage generation in the target non-English language. Experimenting on XL-Sum with Mistral-7B (100 samples per language), we deactivate 5\% of English-heads for \texttt{es} and \texttt{vi}, and 1\% for \texttt{hi}, \texttt{ja}, and \texttt{th}. The language of the generated summaries is classified using FastText \citep{fasttext}. Figure~\ref{fig:thaidemo} illustrates a Thai summarization example where weakening English-heads shifts the output from English to Thai. Table~\ref{tab:outputlan} shows that deactivating a small subset of English-heads increases correct-language outputs to 100\%. Additionally, improved BERTScore F1 confirms that summary quality remains high after this language shift.

These results suggest that language-specific heads can be leveraged to modulate the model's language bias, providing a simple yet effective solution to the off-target language generation problem. Additionally, these findings indicate an asymmetry in multilingual LLMs, showing that while their understanding capability spans many languages, their generation ability remains disproportionately influenced by high-resource languages.

\begin{table}[t]\small
\centering
\begin{tabular}{clccccc}
\toprule
\multicolumn{1}{l}{} &  & es & vi & hi & ja & th \\ 
\midrule
\multirow{2}{*}{Lan-Acc} & VM & 0.67 & 0.35 & 0.74 & 0.99 & 0.78 \\ \cmidrule(lr){2-7} 
 & \textbf{EH} & 1.00 & 1.00 & 1.00 & 1.00 & 1.00 \\ 
 \midrule
\multirow{2}{*}{Sum-Qua} & VM & 57.41 & 50.21 & 70.19 & 81.48 & 58.99 \\ 
\cmidrule(lr){2-7}
 & \textbf{EH} & 71.70 & 80.27 & 85.59 & 81.54 & 69.07 \\ 
 \bottomrule
\end{tabular}
\caption{Language-accuracy ($\uparrow$) and summarization-quality ($\uparrow$) of Mistral-7B on XL-Sum. Deactivating English-heads yields 100\% target language outputs and significantly improves F1 scores versus the vanilla model. \textit{Lan-Acc: Language Accuracy; Sum-Qua: Summarization Quality; VM: Vanilla Model; EH: English-Heads Deactivated.}}
\label{tab:outputlan}
\end{table}

\subsection{Enhancing Multilingual Performance via Language-Specific Head Mask}
Language heads can be leveraged to enhance the multilingual capabilities of LLMs. For a given LLM with $n_l$ layers and $n_h$ attention heads per layer, we construct a trainable matrix of shape $(n_l, n_h)$. Only the parameters corresponding to the identified language heads are set as trainable, while the rest remain frozen, which we refer to as the \textbf{language-specific head mask}. During training or inference, the mask parameters are multiplied with the attention outputs (prior to projection through $W_O$), thereby influencing the model’s final predictions.

Based on each language’s attention head importance matrix, we select the top 2\% scoring heads and set the corresponding positions in the language-specific head mask as trainable parameters. To support language task performance, language-general heads are not excluded from this set. To assess its multilingual effectiveness, we evaluate the language-specific head mask on three models using the XQuAD across languages. We train the language-specific head masks for two epochs on 200 training samples and evaluate on 800 test samples. As a baseline, we train head masks by randomly selecting an equal number of parameters. As shown in Table~\ref{tab:ft_results}, the language-specific head masks yield an average accuracy improvement of 5 percentage points over the vanilla models, and outperform the random baseline by 4 points. These results demonstrate the effectiveness of the language heads identified using LAHIS, and show that updating only a small number of parameters within the language-specific head mask (e.g., 14–20 for these three LLMs), requiring only 30 seconds, can improve multilingual task performance.

\begin{table}[t]\small
\centering
\begin{tabular}{lcccccc}
\toprule
\multicolumn{7}{c}{Aya-23-8B} \\ 
\midrule
 & en & zh & vi & es & el & \textbf{Avg} \\ 
\cmidrule(lr){2-7}
VM & 76.00 & 57.63 & 46.75 & 52.13 & 43.88 & 55.28 \\
RH & 75.25 & 60.75 & 48.00 & 52.88 & 43.88 & 56.15 \\
\textbf{LH} & 77.38 & 63.63 & 53.88 & 60.38 & 50.25 & \textbf{61.10} \\ 
\midrule
\multicolumn{7}{c}{Llama-3.2-3B} \\ 
\midrule
 & en & de & th & vi & el & \textbf{Avg} \\ 
\cmidrule(lr){2-7}
VM & 56.13 & 29.13 & 27.88 & 37.38 & 14.38 & 32.98 \\
RH & 56.50 & 32.00 & 27.00 & 38.25 & 14.63 & 33.68 \\
\textbf{LH} & 59.25 & 35.88 & 31.25 & 39.00 & 18.50 & \textbf{36.78} \\ 
\midrule
\multicolumn{7}{c}{Mistral-7B} \\ 
\midrule
 & en & es & vi & hi & el & \textbf{Avg} \\ 
\cmidrule(lr){2-7}
VM & 44.88 & 30.25 & 21.38 & 9.63 & 6.50 & 22.53 \\
RH & 52.75 & 32.13 & 22.25 & 11.38 & 7.25 & 25.15 \\
\textbf{LH} & 60.13 & 37.13 & 24.13 & 14.88 & 8.88 & \textbf{29.03} \\ 
\bottomrule
\end{tabular}
\caption{Accuracy ($\uparrow$) (\%) on XQuAD. Training language-specific head masks (14–20 parameters) improves accuracy by 5 points over vanilla models and outperforms random masks by 4. \textit{VM: Vanilla Model; RH: Random Head Mask; LH: Language-Specific Head Mask.}}
\label{tab:ft_results}
\end{table}

\section{Related Work}
\textbf{Multilingual Mechanism} Understanding how LLMs internally process different languages remains an open challenge. \citet{llamalan} applies the LogitLens method \citep{logitlens} and finds that in English-centric models like Llama, token representations gradually shift from input space to an English-biased concept space before transitioning to the target language’s output space. \citet{LAPE} proposes a method to identify language-specific neurons, suggesting that a small subset of neurons governs language capability. Similarly, \citet{PLND} introduces the PLND method to detect such neurons and verified a multilingual workflow in LLMs: understanding in multiple language, reasoning in English, and generating output in the target language. \citet{lostml} uses mechanistic interpretability to explore cross-lingual inconsistencies, revealing that most layers encode language-agnostic knowledge, with language-specific transformations occurring mainly in the final layers, where failures often cause output errors. While these works offer valuable insights, they largely focus on entire layers or neurons. In contrast, our work investigates the multilingual roles of attention heads, offering a fresh perspective on cross-lingual behavior in LLMs.

\textbf{Multi-Head Attention} Multi-head attention in LLMs has been shown to be sparse, with only a small subset of heads capturing task-relevant features while others can be pruned without performance degradation \citep{sixteenheads, pruned2019}. However, certain heads have also been demonstrated to play critical roles in many core capabilities of LLMs \citep{conceptsc, multiscalea}. \citet{successorhead} identifies successor heads that increment tokens in natural order, capturing abstract representations shared across architectures. \citet{retrievalhead} discovers retrieval heads responsible for extracting relevant information from long contexts, which are essential for chain-of-thought (CoT) reasoning. \citet{safetyrole} examines the role of attention heads in safety-related behaviors, introducing attribution methods to highlight their influence. Motivated by these findings, our work investigates how certain attention heads contribute to multilingual processing, highlighting a new facet of their functionality.

\section{Conclusion}
This paper provides new insights into the role of MHA in enabling multilingual capabilities in LLMs. We introduce LAHIS, an effective and efficient method for computing attention head importance with respect to multilingual processing. Through comprehensive experiments on three LLMs, we confirm the existence of both language-specific and language-general heads. Our analysis demonstrates that language-specific heads are instrumental in facilitating cross-lingual attention transfer and mitigating off-target language generation, thereby addressing the challenges in multilingual LLMs. Leveraging language heads, we further propose a lightweight adaptation strategy that improves multilingual performance by updating only a small subset of parameters. Collectively, our work not only enhances the interpretability of multilingual behaviors in LLMs from the perspective of MHA, but also presents practical approaches to strengthen multilingual capabilities of LLMs.

\section*{Acknowledgments}
We are grateful to all the anonymous reviewers for their
insightful comments, which highly improve our paper. The research is supported in part by the National Natural Science Foundation of China (No. 62202465) and the National Key Research and Development Program of China (No.2021YFB2910109), and the Outstanding Talent Scheme (Category B) - Qihang Zhou (E3YY141116).

\bibliography{aaai2026}

\clearpage
\appendix
\section{Hardware Setup}
All experiments in this paper are conducted on a single NVIDIA A800 GPU with 80 GB of memory. Furthermore, all LLMs are loaded in the bfloat16 (bf16) format.

\section{Structure of Evaluation Datasets}
We employ the publicly available multilingual datasets XL-Sum and XQuAD in parts of our evaluation. To facilitate the experiments, we apply light preprocessing that restructures the data format without altering its content. Below, we provide examples of the evaluation data and the corresponding prompts used for LLM input. For illustration purposes, all examples are shown in English; in practice, both the data and prompts are rendered in the target evaluation language.

For the XL-Sum dataset, Text~\ref{lst:xlsum-data} and Text~\ref{lst:xlsum-prom} present a sample data entry and its corresponding prompt, respectively.

\lstset{language=Tex, caption={An example from the XL-Sum dataset. It includes the following components: \textbf{text}, the source passage for summarization; and \textbf{summary}, the reference summary provided by the dataset.}, label={lst:xlsum-data}}
\begin{lstlisting}
{
    "text": "By Jon Welch and Paul MoseleyBBC News Details of health problems, family bereavements and personal issues were sent by the University of East Anglia (UEA) in Norwich to 298 students ...",
    "summary": "A university has mistakenly emailed hundreds of students intimate and sensitive personal information about dozens of fellow undergraduates."
}
\end{lstlisting}

\lstset{language=Tex, caption={Zero-shot prompt for XL-Sum. Here, \texttt{\{TEXT\}} is replaced with the \texttt{text} field from the XL-Sum dataset.}, label={lst:xlsum-prom}}
\begin{lstlisting}
Summarize the given text briefly. [Text] {TEXT} [Summary]
\end{lstlisting}

For the XQuAD dataset, Text~\ref{lst:xquad-data} and Text~\ref{lst:xquad-prom} present a sample data entry and its corresponding prompt, respectively.

\lstset{language=Tex, caption={An example from the XQuAD dataset. It includes the following components: \textbf{context}, the background passage used for question answering; \textbf{question}, a query based on the information contained in the context; and \textbf{answer}, the reference answer.}, label={lst:xquad-data}}
\begin{lstlisting}
{
    "context": "The Panthers defense gave up just 308 points, ranking sixth in the league, while also leading the NFL in interceptions with 24 and boasting four Pro Bowl selections. Pro Bowl defensive tackle Kawann Short led the team in sacks with 11, while also forcing three fumbles and recovering ...",
    "question": "Who registered the most sacks on the team this season?",
    "answer": "Kawann Short"
}
\end{lstlisting}


\lstset{language=Tex, caption={Zero-shot prompt for XQuAD. Here, \texttt{\{CONTEXT\}} and \texttt{\{QUESTION\}} are replaced with the \texttt{context} and \texttt{question} fields from the XQuAD dataset.}, label={lst:xquad-prom}}
\begin{lstlisting}
[System Prompt] Based on the given context, give the answer to the question directly. [Context] {CONTEXT} [Question] {QUESTION} [Answer]
\end{lstlisting}

\section{Details of the Cross-Lingual Attention Transfer Experiment}
In \S 3.5, to examine cross-lingual attention transfer facilitated by language-specific heads, we design prompts containing conflicting descriptions of Alice’s profession in two languages, \texttt{Context1} in language A and \texttt{Context2} in language B, alongside an English question.

Specifically, for each selected language, profession contexts are generated by substituting the \texttt{\{OCCUPATION\}} in Text~\ref{lst:context} with each occupation from Text~\ref{lst:pos}. These contexts are then inserted into \texttt{\{CONTEXT1 IN LANGUAGE-A\}} and \texttt{\{CONTEXT2 IN LANGUAGE-B\}} in the prompt template shown in Text~\ref{lst:ta_prom}, ensuring that the two contexts are always in different languages. This procedure yields a diverse set of prompts containing cross-lingual occupational contradictions for evaluation. Examples are displayed in English for demonstration purposes; nevertheless, the actual evaluation uses them in the relevant languages.

\lstset{language=Tex, caption={Context template describing Alice's occupation. Note that no English version is used in actual experiments, as English serves as the neutral language for questioning.}, label={lst:context}}
\begin{lstlisting}
{
	"en": The occupation of Alice is {OCCUPATION}."
}
\end{lstlisting}

\lstset{language=Tex, caption={An example from the occupation list.}, label={lst:pos}}
\begin{lstlisting}
{
"en": ["painter", "scientist", "doctor", "gardener", "lawyer", "dentist", "poet", "writer", "engineer", "director"]
}
\end{lstlisting}

\lstset{language=Tex, caption={Prompt template for the cross-lingual attention transfer experiment, featuring contradictory information in two languages.}, label={lst:ta_prom}}
\begin{lstlisting}
{CONTEXT1 IN LANGUAGE-A} She likes ice-cream. She always do sports in gym at morning. {CONTEXT2 IN LANGUAGE-B} Alice's occupation is a
\end{lstlisting}

Additionally, in this experiment, the magnitude of each attention head is modulated by a gating parameter $g_i$ that allows for its enhancement or weakening:

\begin{equation}
\tilde{head}_i = g_i \cdot head_i, \quad \text{where} \quad g_i \in \mathbb{R}_{\ge 0}
\end{equation}

We conduct evaluations using multiple gate values, specifically \( g_i \in \{0, 2, 3, 5\} \), and report results corresponding to the configuration that achieves the strongest transfer effect.

\section{Language-General Heads Clarification}
In the main text, we demonstrate the existence of language-general heads in multilingual LLMs that significantly contribute to performance across all languages. For rigor, we clarify here that "all languages" refers specifically to the set of evaluated languages in our experiments. However, given that our evaluation spans multiple languages and multiple models, the consistency of the observed pattern suggests a degree of generalizability beyond the specific experimental configurations.

In addition, while these language-general heads play a substantial role in supporting each language, we also observe that their impact varies across languages. In other words, although these heads are broadly beneficial to all evaluated languages, the degree of their contribution is not uniform.

\end{document}